\documentclass[sigconf]{acmart}
\AtBeginDocument{%
  \providecommand\BibTeX{{%
    \normalfont B\kern-0.5em{\scshape i\kern-0.25em b}\kern-0.8em\TeX}}}
    
\def\*#1{\mathbf{#1}}
\def\~#1{\boldsymbol{#1}}

\usepackage{amsmath}
\usepackage{algorithm}
\usepackage{algpseudocode}
\usepackage{graphicx}
\usepackage{caption}
\usepackage{subcaption}

\copyrightyear{2022}
\acmYear{2022}
\setcopyright{acmcopyright}\acmConference[ICAIF '22]{3rd ACM International
Conference on AI in Finance}{November 2--4, 2022}{New York, NY, USA}
\acmBooktitle{3rd ACM International Conference on AI in Finance (ICAIF '22),
November 2--4, 2022, New York, NY, USA}
\acmPrice{15.00}
\acmDOI{10.1145/3533271.3561691}
\acmISBN{978-1-4503-9376-8/22/10}
\begin{document}

\title{Monotonic Neural Additive Models: Pursuing Regulated Machine Learning Models for Credit Scoring}

\author{Dangxing Chen}
\authornote{Corresponding author}
\email{dangxing.chen@dukekunshan.edu.cn}
\author{Weicheng Ye}
\email{weicheng.ye@dukekunshan.edu.cn}
\affiliation{%
  \institution{Zu Chongzhi Center for Mathematics and Computational Sciences, Duke Kunshan Unniversity}
  \streetaddress{8 Duke Avenue}
  \city{Kunshan}
  \state{Jiangsu}
  \country{China}
  \postcode{215316}
}

\begin{abstract}
  The forecasting of credit default risk has been an active research field for several decades. Historically, logistic regression has been used as a major tool due to its compliance with regulatory requirements: transparency, explainability, and fairness. In recent years, researchers have increasingly used complex and advanced machine learning methods to improve prediction accuracy. Even though a machine learning method could potentially improve the model accuracy, it complicates simple logistic regression, deteriorates explainability, and often violates fairness. In the absence of compliance with regulatory requirements, even highly accurate machine learning methods are unlikely to be accepted by companies for credit scoring. In this paper, we introduce a novel class of monotonic neural additive models, which meet regulatory requirements by simplifying neural network architecture and enforcing monotonicity. By utilizing the special architectural features of the neural additive model, the monotonic neural additive model penalizes monotonicity violations effectively. Consequently, the computational cost of training a monotonic neural additive model is similar to that of training a neural additive model, as a free lunch. We demonstrate through empirical results that our new model is as accurate as black-box fully-connected neural networks, providing a highly accurate and regulated machine learning method. 
  
\end{abstract}

\ccsdesc{Computing methodologies~Neural networks.}

\keywords{neural networks; model explainability; fairness }


\maketitle

\section{Introduction}

 Over the past several decades, credit default risk forecasting has been an important research area. As a standard tool, logistic regression (LR) satisfied three regulatory requirements: transparency, explainability, and fairness. To improve prediction accuracy, advanced machine learning (ML) methods have been developed recently \cite{dumitrescu2022machine,finlay2011multiple,paleologo2010subagging}. Some review papers can be found in \cite{baesens2003benchmarking,lessmann2015benchmarking,yeh2009comparisons}. Machine learning could potentially enhance the performance of logistic regression, but there is increasing public concern about the risk of misusing them without meeting regulatory requirements \cite{carlo2021AI}. For highly regulated industries such as the finance sector, model risk management is essential \cite{OCC2021model}. Just recently, the Consumer Financial Protection Bureau (CFPB) confirmed that anti-discrimination law requires companies to provide a detailed sufficient explanation for denying an application for credit when using ML methods \footnote{https://www.consumerfinance.gov/about-us/newsroom/cfpb-acts-to-protect-the-public-from-black-box-credit-models-using-complex-algorithms/}.  As emphasized by CFPB Director Rohit Chopra, ``Companies are not absolved of their legal responsibilities when they let a black-box model make lending decisions." Researchers are now investigating regulated ML tools in light of the growing regulatory requirements.  \cite{chen2021seven,sudjianto2021designing}.
 
 
 Two directions have been extensively explored by researchers to provide explainability \cite{sudjianto2021designing}. The first direction provides model-agnostic methods to disentangle a trained black-box model. Existing popular methods include locally interpretable model-agnostic explanations (LIME) \cite{ribeiro2016should}, SHapley Additive exPlanations (SHAP) \cite{lundberg2017unified}, and sensitivity-based analysis \cite{horel2018,horel2020significance}. Despite their success, black-box ML models are intrinsically not transparent, compared with LR. An alternative direction simplifies ML architectures to increase transparency \cite{Yang2020enhancing, Yang2021gami, chen2018interpretable, agarwal2021neural, dumitrescu2022machine}. The concepts of transparency and explainability are closely related: a transparent model is generally easy to understand. Special attention is paid to \cite{agarwal2021neural} for a new class of neural additive models (NAMs). NAMs combine the approximation capabilities of neural networks (NNs) and the interpretability of the general additive models (GAMs) by taking a linear combination of NNs which are attached to each individual feature. NAMs can be visualized similarly to LR by decomposing a high-dimensional function into a series of univariate functions. In this work, we rely on NAMs' architecture for credit scoring problems for its transparency and explainability. In addition, a sensitivity-based analysis is then applied to determine the importance of a trained model \cite{horel2018} because of its efficiency. Explainability and interpretability are used interchangeably in the paper.

 Aside from transparency and explainability, fairness is another key concern in finance. Algorithmic fairness must be satisfied for companies whenever they use ML methods \cite{uejio2021re}. In credit scoring, fairness is reflected in monotonic constraints on certain features. We present two kinds of monotonicity in this paper using delinquency information as an example. The delinquency features count the number and duration of past dues. Some companies (companies may record features differently) encode this information in different features: each feature counts the number of past dues in the same period. As an example,  delinquency information can be encoded as late payments within six months and more than six months. The following monotonicity constraints are required: (1) monotonicity of individual features: with more past dues, scores must be lowered; (2) monotonicity of pairwise features: if a payment is late for more than six months, the credit score should be lower than if the payment is late within six months.
 The monotonicity of pairwise features is often neglected in the existing literature. Without satisfying monotonicity, even a highly accurate ML method can be hardly accepted by regulators. Various methods have been proposed to impose monotonicity for ML methods \cite{liu2020certified,milani2016fast,you2017deep}. Among them, \cite{liu2020certified} provides a flexible optimization scheme with an efficient certification method and has been very widely used. Using these motivations, we hope to develop an algorithmic approach to credit scoring that incorporates monotonicity.
 
 Here, we present a novel class of monotonic neural additive models (MNAMs), by extending the successful NAMs \cite{agarwal2021neural} with a flexible certified method \cite{liu2020certified} to impose monotonicity. MNAMs penalize monotonicity violations effectively by utilizing the special architectural features of NAMs, thereby avoiding the complex certification process. In consequence, monotonicity is strictly satisfied, and the computational cost of training MNAMs is similar to that of training NAMs. The resulting MNAMs inherit the properties of NAMs, including transparency, explainability, and approximation capability, and maintains fairness by enforcing monotonicity.

Our method is demonstrated using two datasets. First, we show that MNAMs achieve the same accuracy with highly-complex fully-connected neural networks (FCNNs), and significantly outperforms simple LRs, indicating their strong approximation capability; Second, we observe both monotonicity constraints are constantly violated in standard ML methods, resulting in serious algorithmic unfairness, but MNAMs have successfully regulated monotonicity. We would also like to point out that MNAMs may be able to improve generalization accuracy for unseen datasets, which are not revealed by our experiments. The existing datasets have misleading patterns, which result in misguided training for standard machine learning algorithms. As an example, the marginal probability of default with delinquency features is not always monotonically increasing. Several explanations can be offered for this phenomenon, including the impact of the collection agency as well as selection bias in the sample: for delinquent accounts, companies are likely to sell the account to a collection agency, which employs aggressive tactics to collect payments; a record of default might negatively impact a customer's credit score and their ability to obtain credit-based services, so they are not likely to appear in the sample. Without taking account of these impacts, standard ML methods could naively misclassify an obvious high-risk customer as low-risk; on the other hand, MNAMs could easily make a good judgment with enforced monotonicity, just as humans do.   

The structure of this paper is organized as follows. The neural additive model, certified method, and sensitivity-based feature importance method are reviewed in Section 2. We introduce the new monotonic neural additive model in Section 3. Empirical experiments are illustrated in Section 4. In Section 5, we conclude.

\section{Prerequisites}

Assume we have $\mathcal{D} \times \mathcal{Y}$, where $\mathcal{D}$ is the dataset with $n$ samples and $p$ features and $\mathcal{Y}$ is the corresponding labels in classification. For simplicity, we consider the binary classification such that
\begin{align*}
y_i = \begin{cases}
1, & \text{default}, \\
0, & \text{not default}. 
\end{cases}
\end{align*}
We are interested in the probability of default (PoD) of applicants given their information. Credit scores are then determined based on PoD. We assume that the data generating process takes the form of 
\begin{align*}
y|\*x \sim \text{Bernoulli}(f(\*x))    
\end{align*}
for some continuous function $f$. Machine learning (ML) models are then applied to learn $f$.



\subsection{Neural additive models}
\begin{figure}
  \centering
  \includegraphics[scale=0.4]{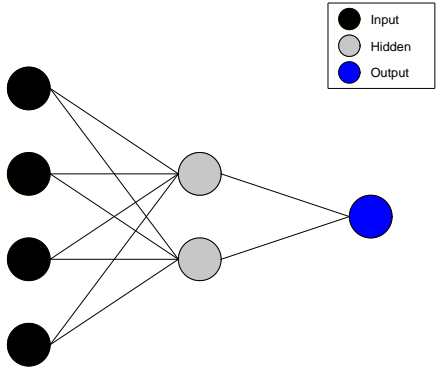}
  \caption{Fully-connected neural network}
  \label{fig:ANN}
\end{figure}

\begin{figure}
  \centering
  \includegraphics[scale=0.4]{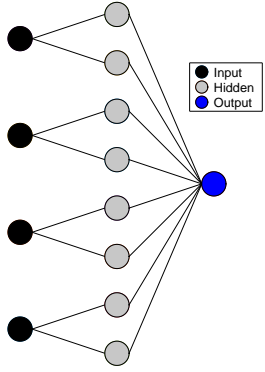}
  \caption{Neural additive model with one hidden layers and two neurons}
  \label{fig:FSNN}
\end{figure}

Fully-connected neural networks (FCNNs) have been very successful for high-dimensional complex functions, due to their universal approximation property \cite{cybenko1989approximation,hornik1991approximation,kubat1999neural,hassoun1995fundamentals}. Despite their success in approximation, their complicated deep layers with massively connected neurons prevent us to interpret the result. Neural additive models (NAMs) \cite{agarwal2021neural} improve the explainability of FCNNs by restricting the architecture of neural networks (NNs). NAMs belong to the family of the generalized additive models (GAMs) of the form
\begin{align*}
g(\mathbb{E}[y|\*x]) = \beta + f_1(x_1) + \dots + f_p(x_p),
\end{align*}
where $\*x = (x_1, \dots, x_p)$ is the input with $p$ features, $y$ is the target variable, $g(\cdot)$ is the link function and is chosen as logistic link function $g(z) = \frac{z}{1-z}$ in our study. For NAMs, each $f_i$ is parametrized by a NN. In an example of four features with one hidden layer and two neurons, the architecture of the NAM in Figure~\ref{fig:ANN} is compared with the FCNN in Figure~\ref{fig:FSNN}.   There are several advantages of NAMs, interested readers are referred to the summary in \cite{agarwal2021neural}. Among all advantages, approximation capability, transparency, and explainability are most essential for credit scoring problems.

\subsection{Certified monotonic neural networks}

In credit scoring, fairness is ensured by imposing monotonic constraints on the output function $f$. We review the certified method in \cite{liu2020certified} to enforce monotonicity. Without loss of generality, assume all monotonic constraints are increasing.  Suppose $\~{\alpha}$ is the list of all monotonic features and $\neg \~{\alpha}$ its complement, then the input $\*x$ can be partitioned into $\*x = (\*x_{\~{\alpha}}, \*x_{\neg \alpha})$. Suppose $\mathcal{X}$, $\mathcal{X}_{\~{\alpha}}$, $\mathcal{X}_{\neg \~{\alpha}}$ are spaces of $\*x, \*x_{\~{\alpha}}, \*x_{\neg \~{\alpha}}$, respectively. Then we have the following definition. 
\begin{definition}
We say $f$ is monotonic with respect to $\*x_{\~{\alpha}}$ if
\begin{align} \label{eq:mono_con1}
& f(\*x_{\~{\alpha}}, \*x_{\neg \~{\alpha}}) \leq f(\*x'_{\~{\alpha}}, \*x_{\neg \~{\alpha}}), \nonumber \\ 
& \forall \*x_{\~{\alpha}} \leq \*x'_{\~{\alpha}},  \forall \*x_{\~{\alpha}}, \*x_{\~{\alpha}}' \in \mathcal{X}_{\alpha}, \forall \*x_{\neg \~{\alpha}} \in \mathcal{X}_{\neg \~{\alpha}},
\end{align}
where $\*x_{\~{\alpha}} \leq \*x_{\~{\alpha}}'$ denotes the inequality for all entries, i.e., $x_{\alpha_i} \leq x_{\alpha_i}'$.
\end{definition}
We say $f$ is individually monotonic on $\*x$ if the following adversarial example does not exist:
\begin{align} \label{eq:adversarial}
f(\widehat{\*x}) > f(\*x),  \text{s.t. } \widehat{\*x}_{\~{\alpha}} \leq \*x_{\~{\alpha}}, \widehat{\*x}_{\neg \~{\alpha}} = \*x_{\neg \~{\alpha}}.
\end{align}
For a differentiable function $f$,  it is verified to be monotonic with respect to $\*x_{\~{\alpha}}$ if 
\begin{align} \label{eq:mono_verify}
\min_{\*x, i} \frac{\partial f (\*x)}{\partial x_{\alpha_i}} \geq 0.    
\end{align}
To accelerate the calculation, Equation \eqref{eq:mono_verify} is reformulated into a  mixed-integer linear programming (MILP) problem, and then solved efficiently by using powerful off-shell techniques. Motivated by \eqref{eq:mono_verify}, to train a monotonic NN, a general Lagrange function is considered
\begin{align} \label{eq:MNN_loss}
\mathcal{L}(\~{\Theta}, \lambda) = \ell(\~{\Theta}) + \lambda h(\~{\Theta}),
\end{align}
where $\ell$ is the typical training loss, $\lambda$ is the Lagrange multiplier, and $h$ is the penalty function that
\begin{align*}
h(\~{\Theta}) = \mathbb{E}_{\*x \sim \text{Uniform}(\mathcal{D})} \left[\sum_{i} \max\left(0,-\frac{\partial f(\*x; \~{\Theta})}{\partial x_{\alpha_i}}  \right)^2 \right].    
\end{align*}
If $h(\~{\Theta}) =0$, then there are no violations, which is equivalent to check \eqref{eq:mono_verify}. In practice, a modified regularization is used by replacing $\max(0,\cdot)$ with $\max(\epsilon,\cdot)$ in $h(\~{\Theta})$. In this paper, we use $\epsilon=10^{-5}$. It is important to check the derivative at all possible points instead of just samples in the dataset, otherwise, adversarial examples \eqref{eq:adversarial} can be found. For high-dimensional inputs, the exact calculation of $h(\~{\Theta})$ becomes intractable due to the curse of dimensionality. As a compromise, a Monte Carlo approach is used with samples of size 1024 that are drawn uniformly from the input domain from iteration to iteration. As violations of monotonicity cannot be exactly tracked by $h(\~{\Theta})$, it is verified through equation \eqref{eq:mono_verify} with an efficient calculation. Starting from $\lambda=0$, the certified algorithm gradually increases $\lambda$ until there are no violations. 


\subsection{Sensitivity-based analysis for feature importance} 

The sensitivity-based method is a convenient way to provide both global and local feature importance \cite{horel2018}. It has main advantage over other approaches for its efficiency. The global feature importance for feature $x_j$ is calculated through
\begin{align} \label{eq:global_fea_import}
\lambda_j = \frac{100}{C} \sqrt{ \frac{1}{n} \sum_{i=1}^n \left( \frac{\partial f(\*x_i)}{\partial x_j} \right)^2 },    
\end{align}
where $C$ is the normalization factor so that $\sum_{j=1}^p \lambda_j = 100$. For simple logistic regressions, $\lambda_j$ reduces to coefficients of features. This metric helps us to rank feature importance by the predictive power of the model, implied by derivative information.

\section{Monotonic neural additive models}

In this section, we utilize the architecture of NAMs and use the certified idea to enforce monotonicity. Before going into details, we discuss two important monotonic constraints in credit scoring, whereas the latter is often neglected in the literature. We use delinquency information as an example. Delinquency features count the number and track the period length of past dues. For some companies, this information is encoded in different features: each feature counts the number of past dues in the same period. Note companies might have different ways to record features, this is only one example. Therefore, two types of monotonicity constraints are required: (1) monotonicity of individual features: more past dues must be penalized with a lower score; (2) monotonicity of pairwise features: with the same number of past dues, the later the customer pays, the worse the impact must be reflected on his/her score. 
In case (2), we need to provide another definition of monotonicity. Analog to \eqref{eq:mono_con1}, we partition $\*x = (x_{\beta},x_{\gamma},\*x_{\neg})$, where $x_{\beta}, x_{\gamma} \in \mathcal{X}_{\beta,\gamma}$ and $\*x_{\neg} \in \mathcal{X}_{\neg}$. Here, $x_{\beta}$ and $x_{\gamma}$ are in the same space so that they can be compared. Then we have the following definition. 
\begin{definition}
We say $f$ is monotonic with respect to $x_{\beta}$ over $x_{\gamma}$ if 
\begin{align}\label{eq:mono_con2}
    & f(x_{\beta},x_{\gamma}+c,\*x_{\neg}) \leq f(x_{\beta}+c,x_{\gamma},\*x_{\neg}), \nonumber \\
    & x_{\beta} = x_{\gamma}, \forall x_{\beta}, x_{\gamma} \in \mathcal{X}_{\beta,\gamma}, \forall \*x_{\neg} \in \mathcal{X}_{\neg}, \forall c \in \mathbb{R}^+. 
\end{align}
\end{definition}
The equation \eqref{eq:mono_con2} leads to the calculation
\begin{align} \label{eq:mono_verify_temp}
    & \frac{f(x_{\beta+c},x_{\gamma},x_{\neg}) -  f(x_{\beta},x_{\gamma},x_{\neg})}{c} \nonumber \\
    -  & \frac{f(x_{\beta},x_{\gamma+c},x_{\neg})- f(x_{\beta},x_{\gamma},x_{\neg})}{c} \geq 0,
\end{align}
which motivates the verification criteria as we discuss later.  Many other features are subject to similar monotonicity, especially when temporal information is involved. For example, hard inquiries, such as those made by a prospective borrower who is actively applying for a credit card or mortgage, should indicate a possible risk associated with the individual. Accordingly, the PoD increases monotonically as the number of inquiries increases. 
It should be noted that in general cases, the condition $x_{\beta}=x_{\gamma}$ must be met. Consider $x_{\beta}$ and $x_{\gamma}$ as the recent and historical inquiries, respectively. For an applicant with no inquiry record, an additional recent inquiry with $(x_{\beta},x_{\gamma}) = (1,0)$ should produce a larger $f$ compared to an additional historical inquiry with $(x_{\beta},x_{\gamma}) = (0,1)$. However, we cannot directly compare the case of $(x_{\beta},x_{\gamma}) = (5,0)$ with the case of $(x_{\beta},x_{\gamma}) = (2,3)$. An applicant with a large $x_{\beta}$ might be affected by the diminishing marginal effect, and $(x_{\beta},x_{\gamma}) = (2,3)$ could also indicate a risky pattern. As far as the case is concerned, there is no clear indication that it carries a high level of risk. 

For a differentiable function $f$, by \eqref{eq:mono_verify} and \eqref{eq:mono_verify_temp}, two monotonic constraints can be verified as follows:
\begin{itemize}
\item Monotonicity of individual features: suppose $\~{\alpha}$ is the list of all individual monotonic features, then
\begin{align} \label{eq:mono_verify1}
\min_{\*x, i} \frac{\partial f}{\partial x_{\alpha_i}}(\*x) \geq 0.    
\end{align}
\item Monotonicity of pairwise features: suppose $f$ is monotonic with respect to $u_i$ over $v_i$ for all i in lists $\*u$ and $\*v$, then
\begin{align} \label{eq:mono_verify2}
\min_{\widetilde{\*x}, i} \frac{\partial f}{\partial x_{u_i}}(\widetilde{\*x}) - \frac{\partial f}{\partial x_{v_i}}(\widetilde{\*x}) \geq 0.    
\end{align}
where $\widetilde{x}_{u_i}=\widetilde{x}_{v_i}$ in $\widetilde{\*x}$. In the example of delinquency features, $u_i$ could denote the number of past dues with a duration period longer than $v_i$.  
\end{itemize}
These conditions can be difficult to check. Fortunately, in NAMs' architecture, conditions \eqref{eq:mono_verify1} and \eqref{eq:mono_verify2} can be simplified as follows.
\begin{itemize}
\item Monotonicity of individual features:
\begin{align} \label{eq:mono1}
\min_i \min_x \frac{\partial f_{\alpha_i}(x)}{\partial x}  \geq 0.
\end{align}
\item Monotonicity of pairwise features:
\begin{align} \label{eq:mono2}
\min_i \min_x \frac{\partial f_{u_i}(x)}{\partial x} - \frac{\partial f_{v_i}(x)}{\partial x} \geq 0.
\end{align}
\end{itemize}
Similar to \eqref{eq:MNN_loss}, we consider the optimization problem:
\begin{align} \label{eq:MNAM_loss}
\min_{\~{\Theta}} \ell(\~{\Theta}) + \lambda h_1(\~{\Theta}) + \eta h_2(\~{\Theta}),
\end{align}
where 
\begin{align*}
h_1(\~{\Theta}) &= \mathbb{E}_{\*x \sim \text{Uniform}(\mathcal{D})} \left[\sum_{i} \max\left(0,-\frac{\partial f(\*x; \~{\Theta})}{\partial x_{\alpha_i}}  \right)^2 \right], \\
h_2(\~{\Theta}) &= \sum_i \widetilde{\mathbb{E}}_{\widetilde{\*x}_i \sim \text{Uniform}(\mathcal{D})} \left[ \max\left(0,-\frac{\partial f (\widetilde{\*x}_i;\Theta)}{\partial x_{u_i}} + \frac{\partial f (\widetilde{\*x}_i;\Theta)}{\partial x_{v_i}}  \right)^2 \right],
\end{align*}
and $\widetilde{\mathbb{E}}_{\widetilde{\*x}_i}$ means that all $\widetilde{x}_j$ except $\widetilde{x}_{v_i}$ are generated from the uniform distribution and $\widetilde{x}_{v_i} = \widetilde{x}_{u_i}$ in $\widetilde{\*x}_i$. 
In practice, as each $f_i$ is a univariate function, we avoid expectation calculation, and it is sufficient to use equispaced $x_j$ in each domain to calculate 
\begin{align*}
h_1(\~{\Theta}) &= \sum_j \sum_{i} \max \left(0, -\frac{\partial f_{\alpha_i} (x_j; \Theta_{\alpha_i})}{\partial x}   \right)^2, \\
h_2(\~{\Theta}) &=  \sum_j \sum_{i} \max \left( 0, -\frac{\partial f_{u_i} (x_j;\Theta_{u_i})}{\partial x} + \frac{\partial f_{v_i} (x_j;\Theta_{v_i})}{\partial x} \right)^2,
\end{align*}
provided $f_i$ is sufficiently smooth. This can be seen as discretization of integral by Riemann sum, which converges with sufficient amount of grids.  As calculation of $h_1$ and $h_2$ are simple, we calculate them exactly and no longer need to check conditions \eqref{eq:mono_verify1} and \eqref{eq:mono_verify2}, in contrast to the original certified method. In the optimization procedure, we also replace all $\max(0,\cdot)$ with $\max(\epsilon,\cdot)$.

We gradually increase $\lambda$ and $\eta$ until penalty terms vanish. The two-step procedure is summarized in Algorithm~\ref{alg:MNAM}. We refer to the NAM that satisfies all required monotonic constraints \eqref{eq:mono1} and \eqref{eq:mono2} as the monotonic neural additive model (MNAM). The monotonic constraints \eqref{eq:mono_verify1} and \eqref{eq:mono_verify2} can be computational expensive to verify for general NNs. MNAMs utilize the special architecture of NAMs, simplify the calculation of the original certified method, and avoid solving complicated verification problems. As a result, the computation expense is almost the same as NAMs. 

\begin{algorithm}[h]
\caption{Certified monotonic additive model}\label{alg:MNAM}
\begin{algorithmic}[1]
\State \textbf{Initialization}: $\lambda=0$ and $\eta=0$
\State Train a NAM by \eqref{eq:MNAM_loss}
\While{$\min(h_1,h_2)>0$}
\State Increase $\lambda$ if $h_1>0$ and increase $\eta$ if $h_2>0$
\State Retrain the NAM by \eqref{eq:MNAM_loss}.
\EndWhile
\end{algorithmic}
\end{algorithm}

\section{Empirical examples}

While there have been many datasets used in the benchmark work \cite{lessmann2015benchmarking}, most of them are either not publicly available or lack feature names. In our study, we focus on two publicly available datasets for detailed discussion. In both datasets, we use the same setup of neural networks (NNs). We compare different methods for both datasets, including logistic regressions (LRs) with no interactions, fully-connected neural networks (FCNNs), neural additive models (NAMs), and monotonic neural additive models (MNAMs). Architectures are rather simple for credit scoring problems. Based on the model architecture optimization, we observe that one hidden layer of two units with the logistic activation function is sufficient for FCNNs, as well as for each NN in NAMs and MNAMs. The results are consistent with existing literature, therefore we do not consider more complicated architectures. We do not compare MNAMs with other ML models due to simplicity. Readers can find some comparisons of NAMs and FCNNs with other models at \cite{agarwal2021neural}. 

There is no single perfect performance measurement for the credit scoring problem. Classification errors, area under the curve (AUC), and confusion matrices are provided. During this study, it is important to note that accuracy is not our primary concern, but rather, regulatory compliance. Thus, we further verify the conceptual soundness and the fairness of the models, which are key requirements for model risk management \cite{OCC2021model}. Visualization along with domain expertise assists in this process. When a model's accuracy is comparable to black-box ML models, models that are compliant with regulatory standards should be chosen. 

\subsection{Taiwan data}

\subsubsection{Description of data}
We first choose the Taiwan credit score dataset \cite{yeh2009comparisons}.
  Among the total 30,000 observations, 6,639 (22.12$\%$) relate to the cardholders with default payments. This research employs a binary variable, default payment, as the response variable.  The dataset is randomly partitioned into the following sets: $75\%$ training set and $25\%$ test set. In addition, this dataset contains the following 23 features:
\begin{itemize}
\item $x_1$: Amount of the given credit (NT dollar): this includes both the individual consumer's credit and his/her family (supplementary) credit.
\item $x_2$: Gender (1 = male; 2 = female).
\item $x_3$: Education (1 = graduate school; 2 = university; 3 = high school; 0,4,5,6 = others).
\item $x_4$: Martial status (1 = married; 2 = single; 3 = divorce; 0 = others). 
\item $x_5$: Age (years).
\item $x_6 - x_{11}$: History of past payments. These variables track the past monthly payment records (from April to September, 2005) as follows: $x_6$= repayment status in September, 2005; $x_7$= the repayment status in August, 2005; \dots; $x_{11}$= repayment status in April, 2005. The measurement scale for the repayment status is $-2$= No consumption; $-1$= Paid in full; 0=The use of revolving credit; 1=Payment delay for one month; 2= Payment delay for two months; \dots; 8=Payment delay for eight months; 9= Payment delay for nine months and above. 
\item $x_{12}-x_{17}$: Amount of bill statement (NT dollar). $x_{12}$= Amount of bill statement in September, 2005; $x_{13}$=Amount of bill statement in August, 2005; \dots; $x_{17}$= Amount of bill statement in April, 2005. 
\item $x_{18}-x_{23}$: Amount of previous payment (NT dollar). $x_{18}$ = Amount paid in September, 2005; $x_{19}$ = Amount paid in August, 2005; $x_{23}$ = Amount paid in April, 2005. 
\item $y$: Client's behavior; 0 = Not default; 1 = Default.
\end{itemize}

\subsubsection{Model performance}

We first apply FCNN to the dataset with and without features $x_2-x_5$. The corresponding AUCs are $78.6\%$ and $78.9\%$, indicating that these features are not important. Therefore, we omit them to avoid potential discrimination. Then, we apply LR, NAM, and MNAM to the dataset. Model performance is summarized in Table~\ref{tab:taiwan_result} and confusion matrices are compared in Table~\ref{tab:taiwan_confusion}.  First, we observe that the FCNN outperforms the traditionally used LR with a $5.3\%$ improvement of AUC, which suggests that ML is a promising tool for improving accuracy. These results are consistent with \cite{yeh2009comparisons}.  Secondly, the NAM performs similarly to the FCNN. With similar accuracy, the NAM should be preferred for its transparency and explainability.

\subsubsection{Conceptual soundness and fairness}

\begin{figure}
    \centering
    \includegraphics[scale=0.4]{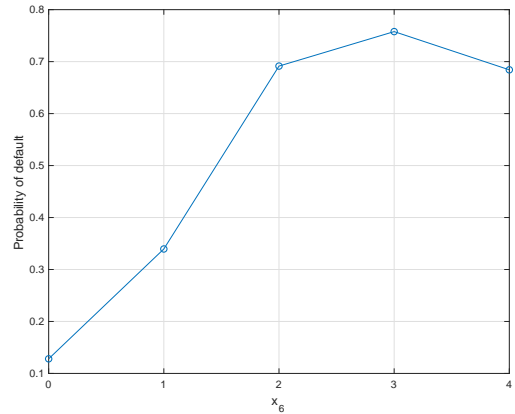}
    \caption{Marginal probability of default with respect to $x_6$ for the Taiwan dataset. A non-monotonic pattern is observed.}
    \label{fig:taiwan_x6}
\end{figure}

Despite the success of FCNN and NAM, monotonicity can easily be violated during the training process. What is worse, patterns in datasets can sometimes be misleading. As an example, we will illustrate this with an analysis of feature $x_6$, the repayment status in September, which is the most important feature, as we will demonstrate later. As we are unsure what the relationship should be between $x_6=-2, -1$, and $0$, we neglect these cases. In a monotonic relationship, it is expected that a larger $x_6$ corresponds to a greater probability of default (PoD). We focus on $x_6=0,1,2,3,4$, as there are only 65 samples for $x_6>4$. To verify monotonicity, we calculate the marginal probability of default with respect to $x_6$, i.e., $\mathbb{P}(y|x_6)$, and plot the result in Figure~\ref{fig:taiwan_x6}. The result is striking: while PoD is monotonic increasing from 0 to 3, it decreases from 3 to 4. It appears puzzling at first glance. Should the PoD decrease for large $x_6$? We argue that Figure~\ref{fig:taiwan_x6} does not reflect the true PoD. There may be several reasons for this, including the effect of the collection agency's intervention and sample selection bias: for late payments, companies are likely to sell the debt to collection agencies, which are known for aggressive methods of collection; a record of default negatively impacts customers' credit scores and their ability to be approved for credit-based services, thereby excluding low credit score customers. Our discussion does not expand upon the detailed description, since it is not contained in the dataset. Nevertheless, a credit assessment ML model should produce a monotonic curve for delinquency features for generalized accuracy and fairness, despite patterns in datasets. 

\begin{figure}
    \centering
    \includegraphics[scale=0.4]{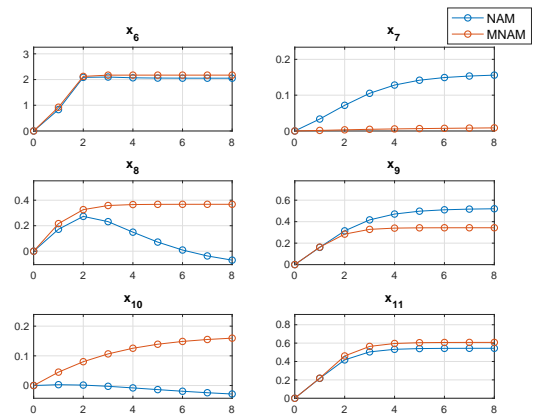}
    \caption{$f_i(x)$ for $i=6, \dots, 11$ of the NAM and the MNAM for the Taiwan dataset. Monotonicities are violated in the NAM.}
    \label{fig:taiwan_x6_x11}
\end{figure}

The observed seeming non-monotonic pattern in Figure~\ref{fig:taiwan_x6} could potentially misguide the unstructured ML methods, and lead to algorithmic unfairness. We focus on the analysis of the NAM, as it performs similarly to the FCNN and is easier to visualize results. For better visualization, we exclude intercepts in all figures here and below since they are not significant for NNs. As shown in Figure~\ref{fig:taiwan_x6_x11}, we plot curves for $f_i(x)$ for $i=6, \dots, 11$, which correspond to all delinquency characteristics. Impacted by the misleading pattern observed in Figure~\ref{fig:taiwan_x6}, $f_6(x)$ decreases for $x>2$. Even worse, $f_8(x)$ and $f_{10}(x)$ have sharper decays for large $x$, which are clearly problematic. In light of these results, NAMs should not be directly applied and monotonic constraints must be imposed.  

Then, we apply MNAM with constraints on $x_6-x_{11}$ to the dataset. The accuracy of the MNAM is comparable to the FCNN and NAM, but monotonicity is imposed successfully. Results are visualized of $f_i$ in Figure~\ref{fig:taiwan_x6_x11}. As algorithmic unfairness is difficult to quantify, it is often overlooked in the existing literature. Nevertheless, one can only use ML methods if monotonicity is satisfied. Furthermore, even though the accuracy of MNAM in the dataset is not significantly improved, we believe that it will perform better for unobserved data. In the case of a customer having a large amount of $x_8$, the NAM would most likely underestimate the risk. As a result, the NAM fails to pass the model stress testing for large values of $x_8$ and $x_{10}$ \cite{OCC2021model}.

\begin{figure}
    \centering
    \includegraphics[scale=0.4]{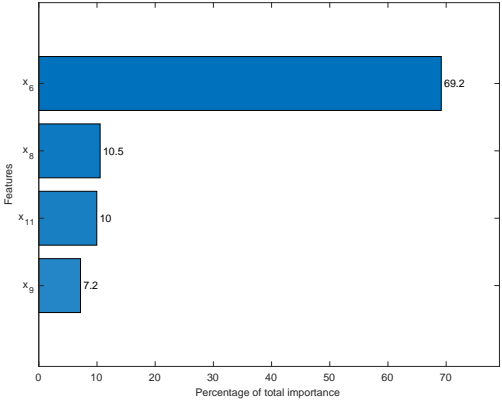}
    \caption{Global feature importance of the MNAM for the Taiwan dataset.}
    \label{fig:taiwan_fea_import}
\end{figure}

\begin{figure}
    \centering
    \includegraphics[scale=0.4]{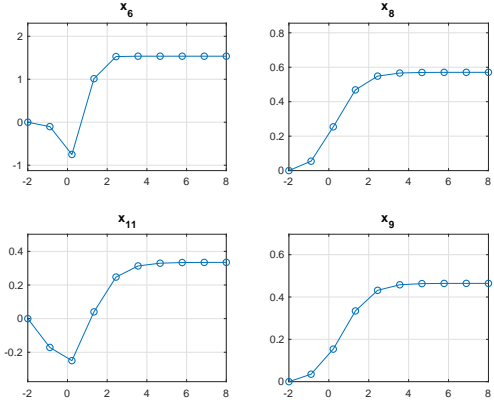}
    \caption{Visualization of NN functions of selected features of the MNAM for the Taiwan dataset.}
    \label{fig:taiwan_fn_visual}
\end{figure}

Having confidence in the MNAM, we then apply the sensitivity-based feature importance method. We calculate feature importance according to Equation~\ref{eq:global_fea_import}. Among the 19 variables, we selected four top features that explain $90\%$ of the significance of the overall features. As shown in Figure~\ref{fig:taiwan_fea_import}, selected features are $x_{6}, x_{8}, x_{11}$, and $x_9$. In the MNAM, each individual NN function corresponding to these selected features can be visualized using Figure~\ref{fig:taiwan_fn_visual}, allowing the user to obtain an intuitive understanding. Visually, we observe an unreasonable pattern of conceptual soundness for delinquency features. For example, although $f$ is monotonic and non-decreasing for $x_6$, it remains almost the same for $x_6>2$. This is undoubtedly incorrect in practice and is misled by the dataset as discussed earlier. This may or may not be a big concern: if $x_6=2$ has already given a strong risky signal to companies, this misbehavior can be safely neglected; otherwise, further work should be done, such as putting additional constraints. Since the solution to this problem is highly dependent on the companies' risk appetite and choice, we do not expand the discussion here. Yet, we wish to point out that such a visualization is not possible with black-box ML methods such as FCNNs.


\begin{table}[h]
    \centering
    \begin{tabular}{ccc}
    \hline
    Model/Metrics  & Classification error & AUC  \\ \hline
    LR & $18.6\%$ & $73.6\%$ \\ \hline
    FCNN & $17.5\%$ & $78.9\%$ \\ \hline
    NAM & $17.4\%$ & $78.9\%$ \\ \hline
    MNAM & $17.2\%$ & $79.2\%$ \\ \hline
    \end{tabular}
    \caption{Model Performance of the Taiwan dataset.}
    \label{tab:taiwan_result}
\end{table}

\begin{table}[h]
    \centering
    \begin{tabular}{ccc}
    \hline
    LR  & Predicted: Default & Predicted: Not default  \\ \hline
    Actual: Default  & 368  & 1255  \\ \hline
    Actual: Not default  & 140 & 5737  \\ \hline
    FCNN  & Predicted: Default & Predicted: Not default  \\ \hline
    Actual: Default  & 593  & 1030  \\ \hline
    Actual: Not default  & 285 & 5592  \\ \hline
    NAM  & Predicted: Default & Predicted: Not default  \\ \hline
    Actual: Default  & 580  & 1043  \\ \hline
    Actual: Not default  & 262 & 5615  \\ \hline
    MNAM  & Predicted: Default & Predicted: Not default  \\ \hline
    Actual: Default  & 594  & 1029  \\ \hline
    Actual: Not default  & 258 & 5619  \\ \hline
    \end{tabular}
    \caption{Confusion matrices of the Taiwan dataset.}
    \label{tab:taiwan_confusion}
\end{table}

\subsection{Kaggle dataset---``Give me some credit"}

\subsubsection{Data description}

Another popularly used dataset is the Kaggle credit score dataset \footnote{https://www.kaggle.com/c/GiveMeSomeCredit/overview}. For simplicity, data with missing variables are removed. Past dues greater than eight times are truncated. Further careful data cleanings could potentially improve model performance but is not the primary concern of this paper. Among the total 120969 observations, 8,357 (6.95$\%$) relate to the cardholders with default payments. This shows that the data are seriously imbalanced. Similarly, the dataset is randomly partitioned into $75\%$ training and $25\%$ test sets. The dataset contains 10 features as explanatory variables:
\begin{itemize}
\item $x_1$: Total balance on credit cards and personal lines of credit except for real estate and no installment debt such as car loans divided by the sum of credit limits (percentage).
\item $x_2$: Age of borrower in years (integer).
\item $x_3$: Number of times borrower has been 30-59 days past due but no worse in the last 2 years (integer).
\item $x_4$: Monthly debt payments, alimony, and living costs divided by monthly gross income (percentage). 
\item $x_5$: Monthly income (real).
\item $x_6$: Number of open loans (installments such as car loan or mortgage) and lines of credit (e.g., credit cards) (integer). 
\item $x_7$: Number of times borrower has been 90 days or more past due (integer). 
\item $x_8$: Number of mortgage and real estate loans including home equity lines of credit (integer). 
\item $x_9$: Number of times borrower has been 60-89 days past due but no worse in the last 2 years (integer). 
\item $x_{10}$: Number of dependents in the family, excluding themselves (spouse, children, etc.) (integer). 
\item $y$: Client's behavior; 1 = Person experienced 90 days past due delinquency or worse.
\end{itemize}

\subsubsection{Model performance}

We first apply FCNN with and without features $x_2$. The corresponding AUCs are $81.1\%$ and $80.0\%$. This suggests the feature age does have some impacts. Nevertheless, age is usually not considered in decision process to avoid discrimination. Hence, we omit it. Then we apply LR, NAM, and MNAM to the dataset. Results are summarized in Table~\ref{tab:GMSC_result} and confusion matrices are compared in Table~\ref{tab:GMSC_confusion}.  The MNAM and NAM achieve the same accuracy as the FCNN and outperform the LR. Based on these results, the MNAM and NAM should be preferred over the LR in terms of their accuracy, and the FCNN in terms of their transparency and explainability.

\subsubsection{Conceptual soundness and fairness}

\begin{figure}
    \centering
    \includegraphics[scale=0.4]{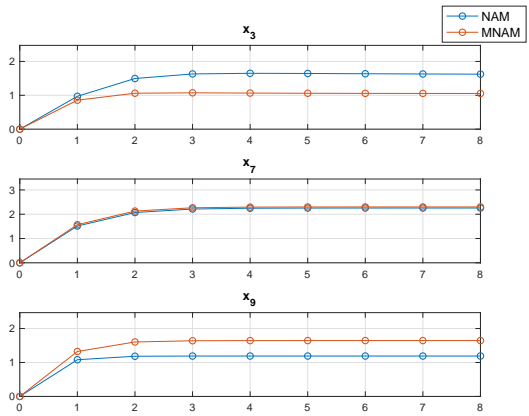}
    \caption{Comparison for $f_3,f_7$ and $f_9$ by the NAM and the MNAM for the GMSC dataset. }
    \label{fig:GMSC_x3x7x9}
\end{figure}

Then we check the algorithmic fairness. We analyze the monotonicity of $x_3, x_7$, and $x_9$ in this dataset, which represent delinquency information. Other monotonic constraints may be easily imposed, but for simplicity, we do not discuss them here. As a starting point, we analyze the monotonicity of individual features. We plot curves of $f_3$, $f_7$, and $f_9$ by the NAM and MNAM in Figure~\ref{fig:GMSC_x3x7x9}. For this training result, the NAM does not seem to violate individual monotonicity.

\begin{figure}
    \centering
    \includegraphics[scale=0.4]{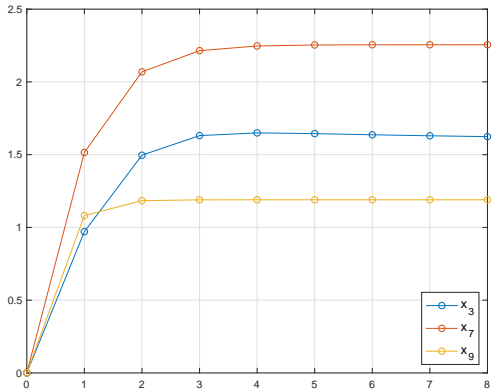}
    \caption{Comparison of $f_3, f_7$ and $f_9$ by the NAM for the GMSC dataset. Monotonicity between $x_3$ and $x_9$ is violated. }
    \label{fig:GMSC_NAM_x3x7x9}
\end{figure}

\begin{figure}
    \centering
    \includegraphics[scale=0.4]{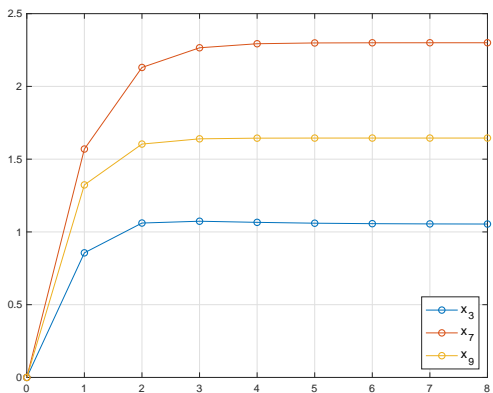}
    \caption{Comparison of $f_3, f_7$ and $f_9$ by the MNAM for the GMSC dataset.}
    \label{fig:GMSC_MNAM_x3x7x9}
\end{figure}

Next, we examine the monotonicity of pairwise features. We determine the relationship that $\frac{\partial f_7}{\partial x}(x)>\frac{\partial f_9}{\partial x}(x)>\frac{\partial f_3}{\partial x}(x)$: a customer who has longer past dues should be considered carrying higher risk to preserve fairness. A comparison of these functions by the NAM is shown in Figure~\ref{fig:GMSC_NAM_x3x7x9}. However, the importance of $x_3$ exceeds that of $x_9$, and monotonicity is violated. To understand intuitively the absurd consequences implied by the NAM, suppose a customer is not paid his past dues within 30 days, he/she should wait for 60-89 days to pay in order to achieve a higher credit score. Then we apply MNAM and show the same plot in Figure~\ref{fig:GMSC_MNAM_x3x7x9}. Monotonicities are successfully achieved by the MNAM and fairness is preserved. 

\begin{figure}
    \centering
    \includegraphics[scale=0.4]{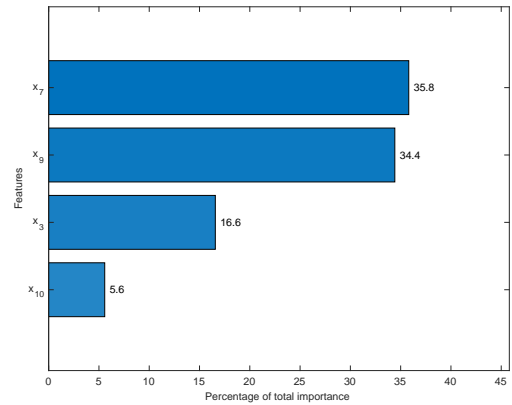}
    \caption{Global feature importance of the MNAM for the GMSC dataset. }
    \label{fig:GMSC_fea_import}
\end{figure}

\begin{figure}
    \centering
    \includegraphics[scale=0.4]{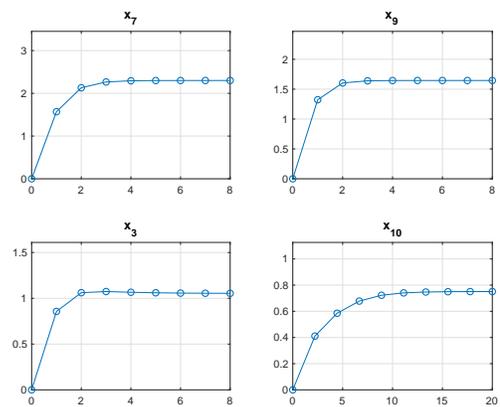}
    \caption{Visualization of NN functions of selected features of the MNAM for the GMSC dataset.}
    \label{fig:GMSC_fn_visual}
\end{figure}

Once the MNAM has been verified, we then apply the sensitivity-based feature importance method. Top features that explain 90 percent of overall features are selected, which represents four of nine variables. As shown in Figure~\ref{fig:GMSC_fea_import}, selected features are $x_{7}, x_{9}, x_{3}$, and $x_{10}$. The result is consistent with our monotonic assumption, demonstrating the usefulness of the sensitivity-based method. The corresponding functions are displayed in Figure~\ref{fig:GMSC_fn_visual}.


\begin{table}[h]
    \centering
    \begin{tabular}{ccc}
    \hline
    Model/Metrics  & Classification error & AUC  \\ \hline
    LR & $6.5\%$ & $77.9\%$ \\ \hline
    FCNN & $6.5\%$ & $80.0\%$ \\ \hline
    NAM & $6.6\%$ & $80.0\%$ \\ \hline
    MNAM & $6.6\%$ & $80.0\%$ \\ \hline
    \end{tabular}
    \caption{Model performance of the GMSC dataset.}
    \label{tab:GMSC_result}
\end{table}

\begin{table}[h]
    \centering
    \begin{tabular}{ccc}
    \hline
    LR  & Predicted: Default & Predicted: Not default  \\ \hline
    Actual: Default  & 272  & 1790  \\ \hline
    Actual: Not default  & 177 & 27828  \\ \hline
    FCNN  & Predicted: Default & Predicted: Not default  \\ \hline
    Actual: Default  & 366 & 1696  \\ \hline
    Actual: Not default  & 253 & 27752  \\ \hline
    NAM  & Predicted: Default & Predicted: Not default  \\ \hline
    Actual: Default  & 344  & 1718  \\ \hline
    Actual: Not default  & 260 & 27745  \\ \hline
    MNAM  & Predicted: Default & Predicted: Not default  \\ \hline
    Actual: Default  & 339  & 1723  \\ \hline
    Actual: Not default  & 257 & 27748  \\ \hline
    \end{tabular}
    \caption{Confusion matrices of the GMSC dataset.}
    \label{tab:GMSC_confusion}
\end{table}

\section{Conclusion}

We present a novel class of monotonic neural additive model (MNAMs) in this paper. As opposed to fully connected neural networks (FCNNs), MNAMs inherit transparency and explainability from state-of-the-art neural additive models (NAMs). To ensure fairness, we also consider two types of monotonicity. The certified method is applied to gradually penalize each individual neural network and is capable of producing a monotonic network efficiently. We obtain accurate, transparent, explainable, and fair MNAMs as a result of this process.

The results of empirical experiments indicate that a simplified structure based on NAMs achieves the same level of accuracy as FCNNs. Because of the transparency of NAMs' structure, we are now able to easily understand and interpret the outputs of neural networks, providing in-depth explanations. Furthermore, we demonstrate that directly applying neural networks to datasets could easily lead to unfair models, particularly in the case of datasets that contain misleading patterns. MNAMs, on the other hand, have produced a fair model, providing more reasonable results. In particular, the current focus in academia and industry has been on the individual monotonicity, but the success of pairwise monotonicity suggests more relationships between different features may be of interest to investigate, and additional restrictions may be necessary to achieve a more reasonable model. In the case of delinquency information, for example, if the company recorded past dues with various periods, as in the Kaggle dataset, then delinquency would exhibit a higher degree of monotonicity, i.e. any changes of longer periods would be more significant than the same amount of changes of shorter period. The purpose of this paper has been to provide a general definition of pairwise monotonicity that applies to many features. It is worth investigating more detailed restrictions by digging deeper into the dataset.

There are several directions we would like to pursue in the future. (1) Our study relies on the architecture of NAMs to achieve monotonicity. It would be possible to extend similar concepts to a flexible architecture that allows for interaction between components. (2) The finance industry is highly regulated, so there are a number of restrictions that must be adhered to. The focus of this paper has been on individual and pairwise monotonicity inspired by fairness, but additional model restrictions may be imposed in order to meet several requirements \cite{repetto2022multicriteria}.  (3) Furthermore, we concentrate on the application of credit scoring due to its strong regulatory requirements. Many other areas of finance are also regulated, such as fraud detection. Regulation requirements may vary. We would like to extend regulated machine learning methods to other areas as well. Overall, we believe that MNAMs will broaden the application of regulated models within the field of machine learning in finance.

\bibliographystyle{ACM-Reference-Format}
\bibliography{sample-base}

\end{document}